%% file: key_point_analysis_emnlp2020.tex
\documentclass[11pt,a4paper]{article}
\pdfoutput=1
\usepackage[hyperref]{emnlp2020}
\usepackage{times}
\usepackage{latexsym}
\usepackage{array,multirow,graphicx}

\usepackage[draft]{todonotes}   % notes showed
\usepackage{amsmath}
\DeclareMathOperator*{\argmax}{argmax}
\usepackage{algorithm} 
\usepackage{algpseudocode} 
\usepackage{wrapfig}
\usepackage{enumitem}

\usepackage{microtype}
\aclfinalcopy % Uncomment this line for the final submission
\def\aclpaperid{2688} %  Enter the acl Paper ID here

\makeatletter
\def\blfootnote{\gdef\@thefnmark{}\@footnotetext}
\makeatother
\title{Quantitative Argument Summarization and Beyond:\\Cross-Domain Key Point Analysis}
\author{ Roy Bar-Haim \qquad Yoav Kantor$^{*}$ \qquad Lilach Eden \qquad Roni Friedman\\ 
 \textbf{Dan Lahav \qquad Noam Slonim}\\
				 IBM Research\\
\texttt{\{roybar,yoavka,lilache,roni.friedman-melamed,noams\}@il.ibm.com}\\
				\texttt{dan.lahav@ibm.com}}

\date{}

\begin{document}
\maketitle
\begin{abstract}
\input{abstract}
\end{abstract}
\section{Introduction}
\input{introduction}
\section{Matching Comments to Key Points}
\input{mapping}
\section{Key Point Extraction}
\input{extraction}
\section{Experiments}
\input{experiments}
%   
\section{Related Work}
\input{related_work}

\section{Conclusion}
\input{conclusion}
\bibliographystyle{acl_natbib}
\bibliography{key_points}
\input{appendix}

\end{document}

%% file: abstract.tex
When summarizing a collection of views, arguments or opinions on some topic, it is often desirable not only to extract the most salient points, but also to quantify their prevalence. Work on  multi-document summarization has traditionally focused on creating textual summaries, which lack this quantitative aspect. Recent work has proposed to summarize arguments by mapping them to a small set of expert-generated \emph{key points}, where the salience of each key point corresponds to the number of its matching arguments. The current work advances \emph{key point analysis} in two important respects: first, we develop a method for automatic extraction of key points, which enables fully automatic analysis, and is shown to achieve performance comparable to a human expert. Second, we demonstrate that the applicability of key point analysis goes well beyond argumentation data. Using models trained on publicly available argumentation datasets, we achieve promising results in two additional domains: municipal surveys and user reviews. An additional contribution is an in-depth evaluation of argument-to-key point matching models, where we substantially outperform previous results. 

%% file: introduction.tex
\blfootnote{
    \hspace{-0.2cm}
		$^{*}$First two authors equally contributed to this work.
}
The need for summarizing views, arguments and opinions on a given topic is common to many text analytics applications, across a variety of domains. Some prominent examples for this type of data are responses to open-ended questions in surveys, user reviews on products and services, and posts in online discussion forums. We will hereafter refer to such utterances that express an opinion, view, argument, ask, or suggestion, collectively as \emph{comments}.

Compressing such textual collections into short summaries relies on their inherent \emph{redundancy}. The goal of Multi-Document Summarization (MDS) algorithms is to create short textual summaries from document clusters sharing the same topic. These summaries aim to capture most of the relevant information in the input clusters, while removing redundancies.  However, in many cases we would also like to \emph{quantify} the prevalence of each of the  points included in the summary. For example, when analyzing the responses of a municipal survey, it would be desirable to let the policy makers know that the point \emph{``The city needs better public transportation''} in the summary matches 8\% of the comments, while the points \emph{``Please consider increasing the number of parks, walking and biking trails.''} and \emph{``electric rates are too high''}  match 4\% and 2\% of the comments, respectively. The users may also want to drill down to view the comments that were mapped to a specific point in the summary.

Recently, \citet{Barhaim:2020} proposed \emph{key point analysis} as a summarization framework that meets the above desiderata, in the context of argument summarization. Given a collection of arguments on some topic, their approach aims to match each argument to a short list of \emph{key points}, defined as high-level arguments. In their work, key points were manually composed by an expert, while the matching of arguments to key points was done automatically. 

The current work promotes this line of research in two important respects. First, we develop a method for automatic key point extraction (Section~\ref{sec:extraction}), allowing fully automatic key point analysis. Our method first selects short, high quality comments as \emph{key point candidates}. It then leverages previous work on argument-to-key-point matching to select a subset of the candidates that achieve high coverage of the data. We show that this relatively simple approach for key point extraction achieves results on argumentation data that are on par with human experts. 

The second major contribution of this work is demonstrating the applicability of key point analysis in additional domains beyond argumentation. We report promising results on two datasets: municipal surveys and user reviews. Remarkably, the results are achieved using the same argument matching and argument quality models that were trained on argumentation data, and require only minimal parameter tuning, but no domain-specific labeled data. 

An additional contribution is an extensive comparison of pre-trained Transformer models for argument matching, in terms of both accuracy and run time, which results in substantial improvement over the best results reported by \citeauthor{Barhaim:2020} (Section~\ref{sec:mapping}).

%% file: mapping.tex
\label{sec:mapping}
\begin{table*}[t]
\small
\centering
\begin{tabular}{|l|l|c|c|c|c|}
\hline
Model &  Selection Policy & Accuracy & Precision & Recall & F1 \\ \hline \hline
BERT & \begin{tabular}[c]{@{}l@{}}TH\\ BM\\ BM+TH\end{tabular} & \begin{tabular}[c]{@{}l@{}}0.867\\ 0.879\\ 0.893\end{tabular} & \begin{tabular}[c]{@{}l@{}}0.677\\ 0.705\\ 0.788\end{tabular} & \begin{tabular}[c]{@{}l@{}}0.700\\ 0.716\\ 0.665\end{tabular} & \begin{tabular}[c]{@{}l@{}}0.685\\ 0.710\\ 0.721\end{tabular} \\ \hline
XLNet & \begin{tabular}[c]{@{}l@{}}TH\\ BM\\ BM+TH\end{tabular} & \begin{tabular}[c]{@{}l@{}}0.897\\ 0.894\\ 0.908\end{tabular} & \begin{tabular}[c]{@{}l@{}}0.750\\ 0.743\\ 0.834\end{tabular} & \begin{tabular}[c]{@{}l@{}}0.759\\ 0.751\\ 0.709\end{tabular} & \begin{tabular}[c]{@{}l@{}}0.752\\ 0.747\\ 0.765\end{tabular} \\ \hline
RoBERTa & \begin{tabular}[c]{@{}l@{}}TH\\ BM\\ BM+TH\end{tabular} & \begin{tabular}[c]{@{}l@{}}0.897\\ 0.895\\ 0.913\end{tabular} & \begin{tabular}[c]{@{}l@{}}0.731\\ 0.745\\ 0.849\end{tabular} & \begin{tabular}[c]{@{}l@{}}\textbf{0.803}\\ 0.753\\ 0.711\end{tabular} & \begin{tabular}[c]{@{}l@{}}0.765\\ 0.749\\ 0.773\end{tabular} \\ \hline
ALBERT & \begin{tabular}[c]{@{}l@{}}TH\\ BM\\ BM+TH\end{tabular} & \begin{tabular}[c]{@{}l@{}}0.909\\ 0.908\\ \textbf{0.926}\end{tabular} & \begin{tabular}[c]{@{}l@{}}0.779\\ 0.778\\ \textbf{0.877}\end{tabular} & \begin{tabular}[c]{@{}l@{}}0.794\\ 0.785\\ 0.751\end{tabular} & \begin{tabular}[c]{@{}l@{}}0.784\\ 0.780\\ \textbf{0.809}\end{tabular} \\ \hline
\end{tabular}
\caption{Argument-to-Key Point matching results on the ArgKP dataset.}
\label{tab:four-models-results}
\end{table*}
The goal of key point analysis is to extract key points and to match comments to these key points. As mentioned in the previous section (and will be further detailed in the next section), our key point selection algorithm is also based on matching comments to key points, making it a critical component in our system. 

We build on the work of \citet{Barhaim:2020}, who developed a large-scale labeled dataset for the task of matching arguments to key points. The dataset, termed \emph{ArgKP}, contains about 24K \emph{(argument, key point)} pairs, for 28 controversial topics. Each of the pairs is labeled as matching/non-matching. Given a set of key points for a topic, an argument could be matched to one or more key points, or to none of them. The arguments in this dataset are a subset of a larger dataset, the \emph{IBM-ArgQ-Rank-30kArgs} dataset, which contains 71 topics, with stance and argument quality annotations for each argument \citep{gretz2019largescale}.

 \citeauthor{Barhaim:2020} only experimented with BERT \citep{bert-2019} as a supervised model for argument matching, which they trained on the ArgKP dataset. We aimed to improve their results by testing several more recent transformer-based pretrained models that were shown to substantially outperform BERT on various tasks \citep{GLUE-2018}, and in particular on the related task of Recognising Textual Entailment (RTE). We used the HuggingFace transformers framework and fine-tuned four different models: bert-large-uncased \citep{bert-2019} (BERT), xlnet-large-cased \citep{xlnet-2019} (XLNet), roberta-large \citep{roberta-2019} (RoBERTa) and albert-xxlarge-v1 \citep{albert-2019} (ALBERT).\footnote{\url{https://github.com/huggingface/transformers}} 

We ran 4-fold cross-validation on the ArgKP dataset, where each fold had a train set of 17 topics, development set (dev-set) of 4 topics and test set of 7 topics. The learning rate for each model was tuned based on the final training loss in one of the splits. % ns1 not sure I understand, but perhaps its fine
%we ran training on the first split with many different learning rates and selected the one that had best results on the train-set (smaller final loss) and dev-set. 
This learning rate was then used in all four splits. The selected learning rates were 2e-5 for BERT, 7e-6 for XLNet, 5e-6 for RoBERTa and 1e-5 for ALBERT. For choosing the number of epochs, we trained each model with 3 epochs and 9 epochs and selected the one that performed better on the dev-set. All models were better when trained for 9 epochs, except BERT that was better when trained for 3 epochs. 

The evaluation results for these models with the above parameters are shown in Table~\ref{tab:four-models-results}. First, we ran inference with each model over all the (argument, key point) pairs in the dev-set and test-set. We then evaluated the following  \emph{selection policies} defined by \citeauthor{Barhaim:2020} A selection policy defines how to match an argument to one or more key points, based on the classifier's match score for each key point (kp), and a given threshold $t$: 
\begin{itemize}[topsep=4pt,itemsep=1pt,parsep=4pt]
    \itemsep0.1em
    \item The \emph{threshold (TH)} policy matches the argument to all the kps with match score $>t$.
    \item The \emph{best match (BM)} policy matches the argument to the kp with the highest match score.
    \item The \emph{best match+threshold (BM+TH)} policy matches the argument to the % ns1 the 
    kp with the highest match score, if the match score $>t$.
\end{itemize}
For each fold, we selected the threshold $t$ that maximizes the F1 score over the dev-set.

 %Eventually, we take the mean scores of the four folds. 
  The model that achieves the best F1 score is ALBERT with an F1 score of 0.809. RoBERTa is second best with an F1 score of 0.773. However, inference time of RoBERTA is about 6 times faster than ALBERT (run times for each model are detailed in Appendix~A). Taking run time into account, we decided, for practical reasons, to use the RoBERTa model in the rest of the experiments. We apply this model to arguments, as well as other types of comments in different domains. Notably, both ALBERT and RoBERTa substantially outperform BERT, which only reaches F1 score of 0.721 (similar to the F1 of 0.713, reported for BERT by \citeauthor{Barhaim:2020}).

%% file: extraction.tex
%(Lilach)
%0.75 page
\label{sec:extraction}
In addition to the matching of comments to given key points, we wish to extract the key points automatically from the set of comments, to enable fully-automatic key point analysis. 
Extraction is performed in two steps: first, a set of key point candidates is selected from the comments and second, the most salient candidates are selected as key points.
\subsection{Candidate Extraction} 
Our approach assumes that the desired key points can be found among the given comments. We start by collecting concise, high quality candidates. We consider only single sentences, and filter out sentences whose length exceed a certain number of tokens. In order to ensure the high quality and argumentative nature of the selected comments, %high comment quality, 
we use 
%a portion of 
the publicly available \emph{IBM-ArgQ-Rank-30kArgs} dataset of \citet{gretz2019largescale}, which consists of around 30k arguments annotated for point-wise quality to train an argument quality ranking model. 

We then use this model to compute the argument quality score of each comment, and include only high quality candidates. %Despite being trained on argumentation data, we found the same model highly effective for obtaining sentences of argumentative nature in other domains as well, when adjusting the quality threshold to the specific dataset using a development set. 
In addition, we filter out sentences starting with pronouns in order to keep the key points self-contained. %\todo{coref resolution seems needed here, but wasn't accurate enough}
\subsection{Key Point Selection}
\label{sec:selection}
After the set of candidates is extracted, we use the matching model described in Section~\ref{sec:mapping} to obtain a match score between each comment and candidate, and between each pair of candidates. 

First, to achieve high coverage of the selected key points, we match comments to candidates by applying the BM+TH selection policy using the matching model and a threshold $t$, and sort the candidates in descending order according to their coverage, i.e., the number of matched comments. Second, in order to avoid redundancy among the selected key points, we traverse the candidates and remove from the list each candidate whose matching score with a higher-ranked candidate exceeded the %\todo{a predetermined similarity} 
threshold.\footnote{As the match scoring function is not symmetric, we compute the match score in both directions and take the average.} The removed candidates and their matched comments are then matched to the remaining candidates. Finally, the candidates are resorted to form a ranked list of top key points. 

The pseudo-code of the algorithm can be found in Appendix~B. % \ref{algorithm:selection}. 

%% file: experiments.tex
\subsection{Evaluaton Method}
\input{method}
\subsection{Datasets}
\input{test_sets}
\subsection{Experimental Setup}
\input{setup}
%
\subsection{Human Evaluation}
\input{evaluation}
%
\subsection{Results and Discussion}
\input{results}
%
%\subsection{Error Analysis}
%\input{error_analysis}

%% file: method.tex
Let $D$ be a dataset, $T$ the set of topics\footnote{Topics may be debate motions in argumentation data, products in user reviews, etc.} in $D$,
$C_t$ the set of comments
for a topic  $t \in T$, and $K_t$ the set of key points extracted for $t$. Key point analysis finds for each $t \in T$ a set of key points $K_t$ and a mapping from a subset of $C_t$ to $K_t$. We define \emph{precision} as the fraction of mapped comments for which the mapping was correct, and \emph{coverage} as the fraction of mapped comments out of all the comments. 

Our goal is to achieve both high precision and high coverage, however there is typically a tradeoff between the two. This tradeoff can be controlled by setting a threshold on the match score, and applying the BM+TH selection policy to match only a
subset of the comments to the key points. 

We explore this tradeoff by measuring the precision for different levels of coverage. The precision at coverage $c$ is defined as the maximal precision  such that the coverage is at least $c$ (which can be found by searching over possible threshold values). We measure precision at coverage levels of $0.2,0.4...,1.0$.

All the configurations in the following experiments use the matching model that was selected as described in Section~\ref{sec:mapping}, and differ only in the set of key points $K_t$ generated for each of the topics $t \in T$ in the dataset $D$.  

The evaluation of each configuration is performed as follows:
\begin{enumerate}[topsep=4pt,itemsep=1pt,parsep=4pt]
    \item For each $t \in T, c \in C_t$ we map $c$ to its best matching key point $k$ in $K_t$, with matching score $s$.
    \item We randomly select from the dataset 500 comments with uniform distribution over the topics. For each sampled comment, we add the tuple $[(c,k),s]$ to our sample.
    \item The $(c,k)$ pairs are manually labeled as matched/unmatched (cf. Section~\ref{sec:annotation}).
    \item Based on the manual labeling of the sample, we measure precision at coverage levels of $0.2,0.4...,1.0$.
\end{enumerate}    

%% file: test_sets.tex
% (Roni)
% \begin{itemize}
%     \item Describe each of the three datasets
%     \item mention that we take only the first sentence in Austin and why we do that
%     \item describe dev\/test splits
% \end{itemize}

We test our key point analysis method on three datasets: \emph{Arguments}, \emph{Survey} and \emph{Reviews}. 

    \paragraph{Arguments Dataset.} The \emph{IBM-ArgQ-Rank-30kArgs} dataset \cite{gretz2019largescale} contains $30$k arguments actively collected for and against $71$ debatable topics, such as \emph{``Homeopathy brings more harm than good''}\footnote{\url{https://www.research.ibm.com/haifa/dept/vst/debating\_data.shtml}}. Arguments in the dataset have strict length limitations. Each argument is annotated for its stance towards the topic it discusses and for its quality. As previously mentioned, \textit{ArgKP} was created based on part of this dataset ($28/71$ topics).
    
    \paragraph{Survey Dataset.} Open-ended comments provided by respondents to the \emph{Austin Community Survey}, which took place in 2016 and 2017\footnote{\url{https://data.world/cityofaustin/mf9f-kvkk}}. Comments were written in response to the following question: \emph{``If there was ONE thing you could share with the Mayor regarding the City of Austin (any comment, suggestion, etc.), what would it be?''}. These comments are raw and unedited, and sometimes contain a few sentences each. The dataset contains $3,188$ comments. Since over 90\% of the arguments in our training data are single sentences, and in order to avoid sentences with missing context, only the first sentence of each comment was included in our set.\footnote{About 50\% of the comments contain a single sentence, and in many of the multi-sentence comments, the first sentence captures the main point of the comment.} % ns1 - the first sentence of each comment? i.e., 3,188 sentences were analyzed?
    
    \paragraph{Reviews Dataset.} The \textit{Opinosis} dataset \citep{ganesan-etal-2010-opinosis} contains sentences extracted from user reviews on a given topic\footnote{\url{https://github.com/kavgan/opinosis-summarization/blob/master/OpinosisDataset1.0_0.zip}}. Each topic is a combination of product name and review aspect, such as \emph{sound quality of ipod nano}. The dataset contains $51$ topics and $7,086$ review sentences, obtained from Tripadvisor (hotels), Edmunds.com (cars) and Amazon.com (various electronics).  

%% file: setup.tex
\paragraph{Data Splits and Model Training.}
We used the 28 topics of the ArgKP dataset as training set (24 topics) and development set (4 topics) for the comment matching classifier, which used the selected model as described in Section~\ref{sec:mapping}. This model was applied to all three datasets. The remaining 43 topics in the Arguments dataset were used as the test set. Following \citeauthor{Barhaim:2020}, we perform key point analysis per topic+stance, 86 pairs in total.

We trained two versions of the argument quality classifier\footnote{Replicating the $BERT\_FT_{topic}$ model of \citet{gretz2019largescale}.}. One was applied to the Arguments test set, so it was only trained on the 24 training topics, with the 4 development topics serving as a development (dev) set. For the Survey and Reviews datasets, we trained a second model on all the available 71 topics. 

We did not have training data for the Survey and Reviews datasets. However, we split each of them into test/dev sets, and used the dev set for experimentation and manual parameter tuning. The Survey dataset was split into 314 dev and 2,840 test comments (after comments filtering, as described below). The Reviews dataset was split into 10 dev and 41 test topics.

%The development set was then used to tune the matching score threshold: for each matching score predicted over the dev set, we computed the F1 score of the mapping while using this score as the matching threshold and the BM+TH selection policy. 
\paragraph{Filtering and Parameter Tuning.} We applied the following filters to each of the datasets. First, comments with non-ascii characters, less than 10 characters, under 4 or over 30 tokens (excluding punctuation marks) were removed. In the Arguments dataset, we also removed 10\% of the comments that had the lowest quality, as predicted by the argument quality classifier. We did not apply this filter to the other datasets, as we found the quality predictions to be less indicative for their comments. Table ~\ref{tab:data-stats} lists the number of topics and comments in the three datasets, before and after filtering. 
 %Lastly, multi-sentence arguments and arguments starting with a pronoun were filtered out. % ns1 not sure we have room for that, but a table with all dataset stats, before and after the filtering, could have been useful
\begin{table*}[t!]
\begin{center}
\small
\begin{tabular}{|l||p{0.07\textwidth}|p{0.15\textwidth}|p{0.07\textwidth}|p{0.15\textwidth}|p{0.07\textwidth}|p{0.15\textwidth}|}
\hline
\multicolumn{1}{|c||}{} & \multicolumn{2}{c|}{Train} & \multicolumn{2}{c|}{Dev} & \multicolumn{2}{c|}{Test} \\ \hline
Dataset & \# Topics & Comments Before / After filtering & \# Topics & Comments Before / After filtering & \# Topics & Comments Before / After filtering \\ \hline \hline 
Arguments & 24 & 10,324/10,324 & 4 & 1,775/1,599 & 43 & 18,398/16,488 \\ \hline
Survey & - & - & 1 & 314/272 & 1 & 2,840/2,425 \\ \hline
Reviews & - & - & 10 & 1,208/1,094 & 41 & 5,878/4,845 \\ \hline
\end{tabular}
\caption{Number of topics and comments per dataset}
\label{tab:data-stats}
\end{center}
\end{table*}

When selecting key point candidates, we aimed to extract about 20\% of the shorter, higher quality comments. Since the datasets vary in their characteristics, we adjusted the thresholds for each of them using the respective dev-set. % ns1 dev-set. 
We selected candidates of up to 12 tokens in Arguments and Reviews, and 10 for Survey. 
The argument quality thresholds were 0.7, 0.4 and 0.35 for Arguments, Survey and Reviews, respectively. %Lastly, multi-sentence comments and comments starting with a pronoun were filtered out.
%For AQ-score we use 0.7 threshold in Arguments, 0.4 threshold in Surveys and 0.35 threshold in Reviews. 

Finally, the key point selection algorithm requires a matching threshold (parameter $t$ in Section~\ref{sec:selection}). 
We tuned this parameter on the dev set of the Arguments dataset, and selected the threshold that maximized the F1, using the BM+TH selection policy. The best threshold (0.856), was used for both the Arguments and Surveys datasets, where key points were extracted for broad topics. The Reviews dataset, however, required finer granularity, as topics were specific aspects of products\footnote{When using threshold 0.856, around 90\% of the comments for most topics were clustered under a single key point.}. Therefore, its threshold was manually set to 0.999 after a few iterations of running the algorithm on the dev set and reviewing the results.
%
%\paragraph{Configurations}
% \begin{itemize}
%     \item list and name each of the configurations
%     \item Describe the parameters used for each dataset/configuration
% \end{itemize}
%
%
%The evaluated configurations differ from each other in the number of 
%We compare several different configurations over the three datasets. Since ArgKP contains 7 manually written key points per topic-stance, we take a similar number of key points (a little more and a little less) and examine mapping results with top 5 and top 10 key points. 
%We will refer to these configurations as Args-5 and Args-10 respectively. 
%In addition, we asked an expert debater to write 7 key points for 10 randomly selected test topic-stance pairs, and compared the mapping results to them with the automatic key point extraction and mapping.
%Survey test set has about 2840 sentences (in contrast to Arguments where each topic-stance has %The Reviews dataset's gold standard consists of several manually written summaries per topic, of two sentences each. We treat all summary sentences as key point candidates and select top two key points per topic using the KP extraction algorithm. The results are compared to the automatic extraction of top 2 key points over all review sentences.  
%we do the same and automatically extract 2 keypoints for each topic. We also compare our automatically extracted keypoints with manually extracted keypoints by taking all gold (manually written) keypoints and selecting the two that most reviews are mapped to.

%% file: evaluation.tex
\label{sec:annotation}
\textbf{Annotation Process.} Using the Appen crowd labeling platform\footnote{\url{https://appen.com/}}, we annotated pairs of comments and key points for match.  
The instructions stated that \textit{``A key point matches a comment if it captures the gist of the comment, or is directly supported by a point made in the comment''}. 
In addition to this binary choice, there was also an option to indicate that either key point or comment were not clear (which we considered as \textit{no match} in our assessment). Each comment and key point pair was annotated by 7 crowd annotators. 
There were three variants of this task:
\begin{itemize}[topsep=4pt,itemsep=1pt,parsep=4pt]
\item Argumentative data - which presented the topic as the context for each comment and key point pair. It also included an additional question regarding the stance of the comment towards the topic, which we used for quality control.
\item Survey data - which mentioned the general context in which the comments were written (a community survey about the city of Austin).
\item Product review data - which presented product and review aspect as the context for each comment and key point pair.
\end{itemize}
For each variant, examples matching the type of data labeled were offered in the guidelines. % ns1 - include the full guidelines in the sup-mat and indicate that here?

We employed the following measures to ensure the annotations quality:
\begin{itemize}[topsep=4pt,itemsep=1pt,parsep=4pt]
    \item Annotator-$\kappa$ score - a score measuring inter annotator agreement, averaging all pair-wise Cohen's Kappa for a given annotator, for any annotator sharing at least 50 judgements with at least $5$ other annotators, as introduced in \citet{toledo-etal-2019-automatic}. Judgements of annotators with annotator-$\kappa < 0.1$ were ignored.
    \item Selected group of trusted annotators - access to the task was limited to a group of annotators with trusted quality, based on previous tasks that were performed for our team, as in \citet{gretz2019largescale}.
	\item Hidden test questions - for the tasks on argumentative data, stance questions functioned as hidden test questions. As they are based on the \textit{IBM-ArgQ-Rank-30kArgs} dataset, their stance was known. Annotators choosing the wrong stance in more than $15\%$ of their annotations, were ignored.
\end{itemize}

We consider a pair as a match if it was labeled as a match by more than 50\% of the annotators. 

\paragraph{Annotations Consistency.} Fleiss' Kappa for the match question on this task was 0.38. In the Arguments dataset, where stance was also labeled, stance Fleiss' Kappa was 0.86. Both were calculated prior to any filtering performed on the results.

Previous work has shown for a variety of NLP annotation tasks that while individual crowd annotations have lower quality than expert annotations, expert-level annotation quality can be achieved by aggregating over sufficient number of crowd annotations \citep{snow-etal-2008-cheap}. Therefore, crowd annotation quality should be assessed primarily by considering the final, aggregated label.

To this end, we tested the consistency of the labeled results over different sets of annotators as follows: $300$ random comment-key point pairs were selected from the Arguments dataset\footnote{This dataset had the lowest Fleiss kappa of the three - 0.34. Survey dataset kappa was 0.41 and Reviews dataset kappa was 0.37}. Each pair was annotated by 14 different annotators. Annotations for each pair were randomly split to two sets, such that each pair in each set had 7 annotations. After processing each set to produce majority labels, Cohen's Kappa obtained between the pair labels of each set was $0.63$. % ns1 - so far we referred to Fleiss kapa, and here we use Cohen's kappa, which looks a bit odd

%In $47$ pairs ($16\%$ of the sample), label was different. Out of those, $15$ pairs had minimal majority ($0.57\%$) in both sets, while in each set there were additional $12$ pairs with minimal majority, adding up to $60\%$ of each set being with minimal majority. This is while in the entire sample, minimal majority takes up $23-24\%$ of the pairs. 
%
%Common reasons for inconsistent results are: (a) \textit{implicit match}, such as in the argument "television provides \textbf{important information.}" and key point "public television provides \textbf{educational programs}"; (b) \textit{imprecise match}, such as in the argument "embryonic research goes \textbf{against god's will} and shouldn't be furthered" and key point: "stem cell research is \textbf{against nature} and should be stopped".

% kappa  for tasks, 0.38 fleiss on match, match-AvgKappa : 0.37; 0.86 fleiss on stance, stance-AvgKappa : 0.83; none failed on stance kappa, 1 on stance tqs (5 rows); 1 failed on match kappa; 
% sbc match fleiss 0.34, avg 0.33
% austin match fliess 0.41, avg 0.46
% opinosis match fliess 0.38, avg 0.37
% consistency comparison on 14 labelers (0.63 kappa on majority label) ; intuition on the reasons for disagreement
%  3 datasets as Arguments, Survey, and Reviews

%% file: results.tex
\begin{table*}[t!]
\begin{center}
\begin{small}
\begin{tabular}{|c|c||c|c||c|c||c||c|c||}
\cline{3-9}
\multicolumn{2}{c|}{}	&	\multicolumn{7}{c|}{\textbf{Precision}}\\			
\cline{2-9}	
\multicolumn{1}{c|}{}	&\multirow{2}{*}{Dataset}      &	 \multicolumn{2}{c||}{Arguments}       &   \multicolumn{2}{c||}{Arguments}               &	Survey   &\multicolumn{2}{c||}{Reviews}\\
\multicolumn{1}{c|}{}	&             &\multicolumn{2}{c||}{(All)} & \multicolumn{2}{c||}{(Subset)}          &    &\multicolumn{2}{c||}{}         \\ % ns1 - replace All with All Topics, and Subset with Subset of Topics?
\cline{2-9}	

\multicolumn{1}{c|}{}	&\multirow{2}{*}{Configuration}&	Auto      &	 Auto   &	Auto      &	Expert &	Auto & Auto	& Gold\\
\multicolumn{1}{c|}{}	&             &  5 KPs    & 10 KPs  &    7 KPs & 7 KPs & 20 KPs & 2 KPs & 2 KPs \\
	
	\hline
 \parbox[t]{2mm}{\multirow{5}{*}{\rotatebox[origin=c]{90}{\textbf{Coverage}}}} &0.2&0.911&0.933&0.843&0.948&0.873&0.814&0.811\\
 &0.4&0.911&0.932&0.843&0.948&0.824&0.796&0.770\\
&0.6&0.906&0.915&0.837&0.905&0.763&0.731&0.642\\
&0.8&0.854&0.883&0.800&0.808&0.638&0.670&0.544\\
&1.0&0.752&0.792&0.696&0.708&0.514&0.568&0.454\\
	\hline
\end{tabular}	
\caption{Results for the Arguments, Survey, and Reviews datasets. 
%For each configuration, the number of key points (KPs) is indicated.  
\label{tab:results}} 
\end{small}
\end{center}
\end{table*}
The results for the three datasets are summarized in Table~\ref{tab:results}. 
% ns1 in the pdf this is Table-4, and the next table is Table-3... so we mention Table-4 before discussing Table-3... better to switch the numbers, if possible
Fully automatic key point analysis is shown to perform well on the Arguments test set: precision of 0.752 and 0.792 when matching all the comments to 5 and 10 key points, respectively. When matching 60\% of the comments, we achieve precision above 90\%. Table~\ref{tab:sbc} shows an example for key points generated for one of the topic+stance pairs in the Arguments datasets, and their distribution over the comments for that topic and stance.  

We also compared our automatic key point extraction to the approach taken by \cite{Barhaim:2020}, where key points were manually created by a debate expert. Following \citeauthor{Barhaim:2020}, the expert composed 7 key points per topic+stance, based on his domain knowledge, and without being exposed to the comments. A total of 70 key points were composed, for 10 randomly-sampled topic+stance pairs from the test set. Comparing the results for these key points with our automatic results for the same number of key points shows that we were able to achieve similar precision (0.696 vs. 0.708) over all the comments (coverage of 100\%). The precision for coverage of 80\% is also comparable (0.8 vs. 0808). For lower coverage rates, the precision for the manual key points is higher.

To evaluate the similarity between our automatically extracted key points and the ones generated by the human expert, we attempted to match each automatic key point to an associated manually composed key point. Out of the 70 KPs, 10 were classified as \emph{Matching} - the key points are essentially the same; 32 were  \emph{Related} - the key points reflect a similar point or one key point is entailed by the other; 16 were \emph{Remote} - the key points are connected but there exists a distinct change that makes them different in essence, and only 12 were unrelated. These results suggest that the automatic process was able, to a large extent, to mimic the analysis of a human expert. We also found that manually composed key points tend to be more abstract, and in some cases a single manual point matched several more specific automatically extracted points.  
\begin{table}[t!]
\begin{center}
\small
\begin{tabular}{|p{0.7\columnwidth}|r|}
\hline                                  
\textbf{Key Point}                           & \textbf{\%} \\ \hline \hline
People who have three minor offences are unfairly punished.           & 30\%  \\ 
\hline
The three strike law has not proved effective in reducing criminality & 15\% \\
\hline
The three strike law prohibits reform of offenders.                   & 12\% \\                   \hline
Many people could pay long sentences for nonviolent crimes            & 12\% \\                      \hline
The three-strikes law has resulted in overcrowded prisons             & 8\% \\                    \hline
The 3 strikes law doesn't allow judicial discretion in sentencing     & 7\%  \\
\hline
The three-strikes law costs tax payers too much money.                & 6\%  \\                       \hline
The three-strikes law is inequitable and targets men of color.        & 5\% \\                         \hline
The three strike law is too strict for some offenders                 & 5\% \\                        \hline
\end{tabular}
\caption{Top key points and their coverage for the topic \emph{``We should abolish the three-strikes laws''} and \emph{Pro} stance from the Arguments dataset, when generating up to 10 key points using the selection algorithm. After generating the key points list, each of the 267 comments is matched to a key point using the BM selection policy. \label{tab:sbc}}
% ns1 - perhaps mention how many sentences were analyzed in total, for this motion/stance ?
\end{center}
\end{table}
\begin{table*}
\begin{center}
\begin{small}
\begin{tabular}{|p{0.3\textwidth}|r|r|p{0.5\textwidth}|}
\hline
\textbf{Key Point}&\multicolumn{1}{c|}{\textbf{\%}}&\multicolumn{1}{c|}{\textbf{P}}	&\textbf{Top Comments}\\
\hline
\hline
Consider a monorail system to help traffic congestion&	9\%&	0.74&	Need much, much better traffic flow, (example, 183 or 620, Palmer).\\
\cline{4-4}
&&&			Traffic flow is terrible!\\
\hline
Austin needs better public transportation	&8\%	&0.90	&For a progressive city, Austin is lacking in public transportation.\\
\cline{4-4}
&&&				Make improvements to public transportation in north Austin.\\
\hline
Affordability of housing and living in Austin&5\%	&0.85&	Address rapidly increasing cost of living\\
\cline{4-4}
&&&				The cost of living here is insane.\\
\hline
Rising property values and taxes.	&5\%	&0.77	&Reduce property taxes and housing costs so that retiring and still living here is a real possibility.\\
\cline{4-4}
&&&				*This city is not affordable due to horrendous tax and service fees including all city service bills - electric, water, etc. \\
\hline
Please consider increasing the number of parks,walking and biking trails.&4\%&	0.84&	Consider better developed bike lanes throughout the city.\\
\cline{4-4}
&&&				Developing of greenery areas and more parks.\\
\hline
Austin utility services need an overhaul-especially water/wastewater.	&4\%	&0.78	&City needs to fix serious drainage issues, and let citizens protect their homes while they await a cure.\\
\cline{4-4}
&&&				Water/wastewater rates are ridiculous.\\
\hline
\end{tabular}
\caption{Top key points for the City of Austin Community Survey. Match threshold was set so that the extracted 20 key points cover 60\% of the sampled comments. For each key point we show the percentage of matching comments (out of the sampled comments), the precision of matched comments and the top two matching comments. All comments shown in the tables were judged as correct matches, except for the one marked with '*'. \label{tab:austin}} 
\end{small}
\end{center}
\end{table*}

Remarkably, our method, which makes use of models trained on argumentation data, performs reasonably well also when applied to survey and user reviews data. Presumably, the comments in these datasets also contain
argumentation, which allows to transfer the knowledge learned from the argumentation dataset to these domains. The argumentation in the Arguments dataset is more explicit, though, as the contributors to this dataset were asked to provide pro and con arguments for the given controversial topics.

For the Survey dataset, we achieve precision of 0.763 when matching 60\% of the comments in the labeled sample to 20 key points\footnote{We used here a larger number of key points since, unlike the other two datasets, the Survey test set contains a single topic with more than $2,400$ comments.}. Table~\ref{tab:austin} shows KP analysis results for this coverage rate, including the extracted key points, their distribution, comment matching precision per key point, and the top two matching comments for each key point. While the extracted key points are largely concise and to the point, the results could be further improved with some manual post-processing. For example, the first KP can be rephrased as \emph{``Reduce traffic congestion''}, removing the extra part about a monorail system, which is not mentioned in the top comments. We can then re-match the comments to the revised KPs, and the process can iterate, until both coverage and precision are satisfactory. 

The precision over all the comments was 0.514. We note that key point analysis can be effectively applied even if the matching precision is not very high. For example, suppose that 10\% of the comments were matched to a certain key point with precision of only 50\%. This means that in practice, 5\% of the comments do match this key point, so it is an important point nonetheless. 

For the Reviews dataset, we selected two key points per topic, since this was the length of summaries in the experiments conducted by \citeauthor{ganesan-etal-2010-opinosis} on this data. We compared our results to a configuration where the key point candidates are the union of the sentences in the human-generated gold summaries that were released as part of this dataset. 

We obtained precision of 0.731 for coverage of 60\%, and 0.568 for 100\% coverage, better than the results for the key points that were based on the gold summaries (0.642 and 0.454, respectively). The precision differences in coverage levels of 0.6 and above are statistically significant\footnote{Using $Z$ test for two population proportions, with $p=0.05$ for coverage of 0.6, and $p=0.01$ for coverage of 0.8 and 1.0.}. Table~\ref{tab:opinosis} shows both the automatically extracted key points and the key points selected from human summaries, along with their coverage, for several selected topics.

\begin{table*}[]
\small
\begin{tabular}{|l||l|r||l|r|}
\hline 
\multicolumn{1}{|c||}{\textbf{Topic}}                                                                   & \multicolumn{1}{c|}{\textbf{KPs Extracted from Gold Summaries}}                                                                                    & \textbf{\%} & \multicolumn{1}{c|}{\textbf{KPs Extracted from Reviews}}                                                 & \textbf{\%} \\ \hline \hline
\multirow{2}{*}{\textbf{\begin{tabular}[c]{@{}l@{}}Accuracy of \\Garmin nuvi 255W \\GPS\end{tabular}}}   & \begin{tabular}[c]{@{}l@{}}The garmin seems to be generally\\ very accurate.\end{tabular}                                                          & 73\%          & \begin{tabular}[c]{@{}l@{}}Most of the times, this info was very\\ accurate.\end{tabular}                & 72\%          \\ \cline{2-5} 
                                                                                                       & \begin{tabular}[c]{@{}l@{}}Set-up and usage are considered to\\ be very easy.\end{tabular}                                                         & 13\%        & \begin{tabular}[c]{@{}l@{}}Easy to use, excellent accuracy, nice\\ and intuitive interface.\end{tabular} & 16\%          \\ \hline 
\multirow{2}{*}{\textbf{\begin{tabular}[c]{@{}l@{}}Battery-life of\\ iPod Nano 8GB\end{tabular}}}      & \begin{tabular}[c]{@{}l@{}}The battery life of the ipod nano is\\ very short.\end{tabular}                                                         & 79\%          & The only bad thing is it's battery life.                                                                 & 90\%          \\ \cline{2-5} 
                                                                                                       & \begin{tabular}[c]{@{}l@{}}It seems to continue using battery even\\ when the ipod is not in use, otherwise, \\ it's a great product.\end{tabular} & 8\%           & \begin{tabular}[c]{@{}l@{}}Long battery life and easy directions\\ make this a snap to use.\end{tabular} & 7\%           \\ \hline 
\multirow{2}{*}{\textbf{\begin{tabular}[c]{@{}l@{}}Rooms of \\ Bestwestern Hotel \\ SFO\end{tabular}}} &\begin{tabular}[c]{@{}l@{}} Good, clean and tidy rooms and \\bathroom.  \end{tabular}                                                                                                         & 39\%          & \begin{tabular}[c]{@{}l@{}}The hotel had nice, well, decorated,\\ fairly, modern rooms.\end{tabular}     & 30\%          \\ \cline{2-5} & The rooms were small.                            & 25\%          & \begin{tabular}[c]{@{}l@{}}The rooms are a bit small, but not\\ unusual for San Francisco.\end{tabular}  & 18\%          \\ \hline
\end{tabular}
\caption{Top two key points extracted from gold summaries and from original reviews, on selected topics from the Reviews dataset. For each key point we show the percentage of matching comments, with match threshold 0.999.%For gold summaries, the key points were automatically selected from the union of all the sentences in the reference summaries.
\label{tab:opinosis}}
\end{table*}

\paragraph{Error Analysis.} The dominant types of matching errors differed amongst the datasets. The most common type in Reviews data was the comment and KP having opposite  polarity. This was expected, since in the ArgKP training data, the argument and the key point always have the same polarity. The highest proportion of comments that were not related to the key point was in the Survey dataset. This is likely an outcome of analyzing all the comments in this dataset under a single topic. The dominant issue in the Arguments dataset was key points that were related to the comment but had slight changes that altered their meaning. For example: \emph{``if people have been committing crimes anyway, they deserve to be caught through whatever means necessary, including the use of entrapment.''} was matched to \emph{``sometimes the only way police can catch a criminal is by entrapment.''}, where the phrase \emph{``the only way''} was crucial for capturing the right meaning of the key point.

%% file: related_work.tex
The task of \emph{Multi-document summarization (MDS)} is defined as creating a summary that is both concise and comprehensive from multiple % ns1 a cluster of 
texts with a shared theme, e.g., news articles covering the same event \citep{McKeown:2002}. 
%The series of Document Understanding and Text Understanding Conferences (DUC and TAC), organized by NIST, have been the main forums for evaluating MDS systems. 
A major challenge for applying supervised learning to MDS has been the limited amount of available training data. Most of the approaches applied to the task were extractive, operating over graph-based representations of sentences or passages \citep{Erkan:2004,christensen-etal-2013-towards}. Recently, \citet{DBLP:conf/iclr/LiuSPGSKS18} proposed a method for creating a large-scale dataset from Wikipedia (WikiSum), which allowed training an abstractive neural model for this task. Key point analysis adds a quantitative dimension that is not addressed by MDS, by measuring the prevalence of each point in the summary. % ns1 perhaps add - at the cost of generating a bullet-like summary, assuming this is sufficient for various use cases. 

Many of the works on Opinion Summarization take an alternative, sentiment-based approach. These works aim to identify the main aspects discussed in user reviews, and quantify the sentiment towards each of these aspects \citep{hu:kdd04,snyder-barzilay-2007-multiple,titov-mcdonald-2008-joint}. However, as noted by \citet{ganesan-etal-2010-opinosis},  it is still hard for a user to understand why an aspect received a particular rating. As demonstrated in Table~\ref{tab:opinosis}, % ns1 Tabel~\ref{tab:opinosis}, 
key points can address this limitation by providing a more informative summary of user reviews. However, the detection of the stance (or sentiment) of each key point with respect to the topic was left out of the scope of the current work, and we plan to address it in future work. 
% ns1 - but without handling the sentiment... so perhaps here you should refer that targeted sentiment analysis could certainly be valuable on top of kp analysis, and was left out of the scope of this work

In computational argumentation, several works have focused on pairwise argument similarity and clustering  \citep{ajjour-etal-2019-modeling, reimers-etal-2019-classification, misra-etal-2016-measuring}. These works, however, did not attempt to create textual summaries from the resulting clusters.  \citet{egan-etal-2016-summarising} summarized argumentative discussions through the extraction of salient ``points'', where each point is a verb and its syntactic arguments. 
%Applying their unsupervised method to online political debates showed significant improvement over a baseline extractive summarizer, according to human evaluation. 
The current work also extracts points from argumentative data, but our goal is to go beyond textual summaries, by matching each key point to its corresponding sentences in the input data. Similar to \citeauthor{egan-etal-2016-summarising}, we also experimented with extracting syntactic subtrees as key points, but found that this often results in incomplete sentences or omission of important information. Selecting short, high quality sentences as key points was found to perform better in our experiments.

The line of research that is most relevant to the current work deals with matching argumentative texts to predefined, short lists of manually-composed arguments or points \citep{hasan-ng-2014-taking, boltuzic-snajder-2014-back, naderi-2016}.  \citet{Barhaim:2020} matched crowd-contributed arguments, taken from the dataset of \citet{gretz2019largescale}, to key points composed by a debate expert. We used the labeled dataset developed by \citeauthor{Barhaim:2020} to train our comment matching model.

As previously discussed, the main contributions we make to this line of work are (i)~Fully-automatic key point analysis, enabled by automatic key point extraction, and (ii)~Demonstrating the applicability of key point analysis to additional domains besides argumentation, including surveys and user reviews. Furthermore, we were able to achieve promising results on these domains using models that were only trained on argumentation data.

%% file: conclusion.tex
Key Point Analysis is a novel framework for summarizing arguments, opinions and views. It provides both textual and quantitative view of the main points in the summarized data, and allows the user to interactively drill down from points to the actual sentences they cover. Previous work only applied key point analysis in the context of argumentation data, and required a domain expert for writing the key points. 

The current work addresses both of the above limitations. First, we present an automatic method for key point extraction, which is shown to perform on par with a human expert.  Second, our work demonstrates the potential of key point analysis in multiple domains besides argumentation. Furthermore, we show that the necessary knowledge for key point analysis, once acquired by supervised learning from argumentation data, can be successfully applied cross-domain, making it unnecessary to collect domain specific labeled data for each target domain. 

In future work, we would like to improve comment matching, e.g., by making it stance-aware. We also plan to experiment with sequence-to-sequence neural models for generating key point candidates from comments. 

%% file: appendix.tex
\section*{Appendices}
\appendix
\section{Matching Models Run Times}
Table \ref{tab:run-time-measurments} lists run-time measurements for one of the splits of the ArgKP dataset: %The measurements show run-time of one split (split-0). 
 training over 15,235 argument-kp pairs in the train-set and inference over 3,776 pairs in the dev-set and 6,839 pairs in the test-set, using an NVIDIA Tesla V100 GPU. 
 \begin{table}[h!]
\small
\centering
\begin{tabular}{|l||l|l|l|}
\hline 
 & Train & Dev & Test \\ \hline \hline
BERT & 00:18:59 & 00:00:17 & 00:00:32 \\ \hline
XLNet & 01:09:38 & 00:00:23 & 00:00:42 \\ \hline
RoBERTa & 00:59:43 & 00:00:17 & 00:00:31 \\ \hline
ALBERT & 01:06:50 & 00:01:39 & 00:03:03 \\ \hline
\end{tabular}
\caption{Run time (hours:minutes:seconds)}
\label{tab:run-time-measurments}
\end{table}
\section{Key Point Selection Algorithm}
The pseudo-code of the key point selection algorithm (Section 3.2) is listed in Algorithm ~\ref{algorithm:selection}. Given a set of comments, a set of key point candidates and a threshold $t$, the algorithm outputs a sorted list of selected key points.

\begin{algorithm}
\small
\caption{Key Point Selection \newline \textbf{Input}: \small{Comments $C$, KP Candidates $K$, Threshold $t$} \newline
\textbf{Output}: \small{A ranked subset of $K$}
} 
	\begin{algorithmic}[1]
	    \Procedure{Select\_Key\_Points}{$C$, $K$, $t$}
        \State $k\_to\_c \gets  Get\_Matches(C, K, t)$ 
        \State $K \gets $ sort\_descending(keys of k\_to\_c) by \#matches
	    \State $R \gets []$
	    \For{$k1$ in $K$} 
	        \For {$k2$ in $K$ up to and excluding k1} 
	            \State $ s \gets Avg(Score(k1,k2),Score(k2,k1))$
	            \If {$s>t$} % ns1 - is this the same t as in the other procedure?
	                \State add $ k1 \cup k\_to\_c[k1]$ to $R$
	                \State remove $k1$ from $k\_to\_c$
	                \State break
	            \EndIf
	       \EndFor
	   \EndFor
	   \State $kps \gets $ keys of $k\_to\_c$
	   \State $kp\_to\_c \gets k\_to\_c \cup Get\_Matches(R, kps, t)$
	   \State \Return sort\_descending(keys of $kp\_to\_c$) by \#matches
	   \EndProcedure
	   \State
	   \Procedure{Get\_Matches}{$C$, $K$, $t$} 
       \State $k\_to\_c \gets \{\}$ 
       \For{$c$ in $C$}
 	        \State $ match\_c \gets \argmax_{k \in K} Score(c,k)$
 	        \If {$Score(c,match\_c) > t$}
 	            \State add $c$ to $k\_to\_c[match\_c]$
 	        \EndIf
       \EndFor
       \State \Return $k\_to\_c$
 	   \EndProcedure
	\end{algorithmic} 
\label{algorithm:selection}
\end{algorithm}